\documentclass[sigconf,9pt]{acmart}
\usepackage{fancyhdr}
\usepackage{graphicx}
\usepackage{amsmath}
\usepackage{soul}
\usepackage{xcolor}
\usepackage{verbatim}
\usepackage{subcaption}
\usepackage{algorithm}
\usepackage{textcomp}
\usepackage[export]{adjustbox}
\usepackage{siunitx}
\usepackage{array}
\usepackage{commath}
\usepackage{url}
\usepackage{paralist}
\usepackage{booktabs} 
\usepackage{algpseudocode}
\usepackage{gensymb}
\usepackage{enumitem}
\usepackage{cancel}
\usepackage{todonotes}
\usepackage{pdfpages}
\usepackage{animate}
\usepackage{multirow}
\usepackage{attachfile2}

\setlist{nolistsep}
\graphicspath{ {./figs/} }

\newcommand{\argmin}{\arg\!\min}

\def\BibTeX{{\rm B\kern-.05em{\sc i\kern-.025em b}\kern-.08em
    T\kern-.1667em\lower.7ex\hbox{E}\kern-.125emX}}

\newcommand{\tsc}[1]{\textsuperscript{#1}} 

\setcopyright{none}
\renewcommand\footnotetextcopyrightpermission[1]{}
\begin{document}



\title{ORBIT-2: Scaling Exascale Vision Foundation Models for Weather and Climate Downscaling
}


\author{Xiao Wang\tsc{1}, Jong-Youl Choi\tsc{1}, Takuya Kurihaya\tsc{1}, Isaac Lyngaas\tsc{1}, Hong-Jun Yoon\tsc{1}, Xi Xiao\tsc{4}, David Pugmire\tsc{1}, Ming Fan\tsc{1}, Nasik M. Nafi\tsc{1}, Aristeidis Tsaris\tsc{1}, Ashwin M. Aji\tsc{2}, Maliha Hossain\tsc{1} Mohamed Wahib\tsc{3}, Dali Wang\tsc{1}, Peter Thornton\tsc{1}, Prasanna Balaprakash\tsc{1}, Moetasim Ashfaq\tsc{1}, Dan Lu\tsc{1}}

\affiliation{%
  \country{\vskip .2cm} 
  \city{}
  \institution{\tsc{1} Oak Ridge National Laboratory, Oak Ridge, United States}
  \institution{\tsc{2} AMD Research and Advanced Development, Santa Clara, United States}
  \institution{\tsc{3} RIKEN Center for Computational Science, Kobe, Japan}
  \institution{\tsc{4} University of Alabama at Birmingham, United States}
}

\renewcommand{\shortauthors}{X Wang et al.}

\begin{abstract} 
Sparse observations and coarse-resolution climate models limit effective regional decision-making, underscoring the need for robust downscaling. However, existing AI methods struggle with generalization across variables and geographies and are constrained by the quadratic complexity of Vision Transformer (ViT) self-attention. We introduce ORBIT-2, a scalable foundation model for global, hyper-resolution climate downscaling. ORBIT-2 incorporates two key innovations: (1) Residual Slim ViT (Reslim), a lightweight architecture with residual learning and Bayesian regularization for efficient, robust prediction; and (2) TILES, a tile-wise sequence scaling algorithm that reduces self-attention complexity from quadratic to linear, enabling long-sequence processing and massive parallelism. ORBIT-2 scales to 10 billion parameters across 65,536 GPUs, achieving up to 4.1 ExaFLOPS sustained throughput and 74–98\% strong scaling efficiency. It supports downscaling to 0.9 km global resolution and processes sequences up to 4.2 billion tokens. On 7 km resolution benchmarks, ORBIT-2 achieves high  accuracy with $R^2$ scores in range of 0.98–0.99 against observation data.

\end{abstract}

\maketitle

\section{Justification For ACM Gordon Bell Prize for Climate Modelling}
\label{sec:justification}


ORBIT-2 sets new benchmarks in scalability and scientific impact, training ViT-based models with up to 10 billion parameters on 65,536 GPUs with 74–98\% strong scaling efficiency and 4.1 ExaFLOPS sustained throughput in BF16 precision. It enables global climate downscaling at 0.9 km resolution and processes sequences up to 4.2 billion tokens—vastly surpassing prior limits. This breakthrough advances understanding of fine-scale extremes and improves climate modeling critical for adaptation, risk mitigation and decision making.

\begin{center}
\resizebox{\linewidth}{!}{
\begin{tabular}{|c|c|}
\hline
Attributes & Contents \\
\hline
 Category & \begin{tabular}{@{}c@{}}{\em Time-to-solution}, {\em Scalability}, {\em Throughput}\\ \end{tabular} \\
\hline
Type of method & {\em Dense Vision Transformer Model}\\
\hline
Results reported on & {\em Whole application including I/O}\\
\hline
Precision reported & {\em BF16 mixed precision}\\
\hline
System scale & {\em Measured on Full-Scale System} \\
\hline
Measurement mechanism & \begin{tabular}{@{}c@{}}{\em Timer}, {\em Static analysis tool,}\\ {\em FLOP counts}\end{tabular} \\
\hline
\end{tabular}}
\label{Attributes}
\end{center}
\section{Problem Overview}
\label{sec:problem_overview}

Many regions lack dense ground-based observational networks, hindering early warning systems, disaster risk mitigation, and climate adaptation planning. In such cases, global climate models provide a crucial alternative, simulating atmospheric processes at planetary scale. However, their coarse resolution limits the representation of fine-scale phenomena, constraining accuracy at regional levels.

Downscaling bridges this gap by translating coarse resolution global climate model into fine-scale outputs~\cite{Giorgi19}. This process is critical across many sectors, including agriculture~\cite{jones2013generating}, water resources~\cite{gutmann2014intercomparison}, infrastructure planning~\cite{girvetz2013making}, energy systems~\cite{gonzalez2017simulating}, and extreme event forecasting~\cite{hewitson1996climate}. Despite its importance, downscaling remains both scientifically and computationally challenging. It requires physically consistent predictions from massive, high-dimensional spatiotemporal data while maintaining accuracy across diverse regions. These challenges highlight the urgent need for scalable, high-fidelity downscaling approaches to enable effective climate services, disaster preparedness, and policy planning.

Traditional downscaling methods are either dynamical~\cite{giorgi2015regional}, which uses nested physical models to simulate fine-scale processes but is computationally intensive, requiring days or weeks on supercomputers and limited to regional domains; or statistical~\cite{wilby2013statistical, wilby1998statistical}, which is less computationally demanding but often lacks physical fidelity and generalizability. More recently, artificial intelligence (AI) has emerged as a powerful alternative, providing high-resolution predictions at a fraction of the computational cost-even on edge devices~\cite{koldunov2024emerging, kumar2021deep, wang2021deep}. Most existing AI approaches, however, are task-specific deep learning models~\cite{annau2023algorithmic, eyring2024ai, liu2024generative, aich2024conditional} that map coarse to fine resolution, requiring retraining for each variable, resolution, or region, and often struggling to generalize across diverse, physically distinct climate variables~\cite{wang2024orbitoakridgebase}.

\begin{figure*}[t]
\centering
\includegraphics[width=.84\linewidth]{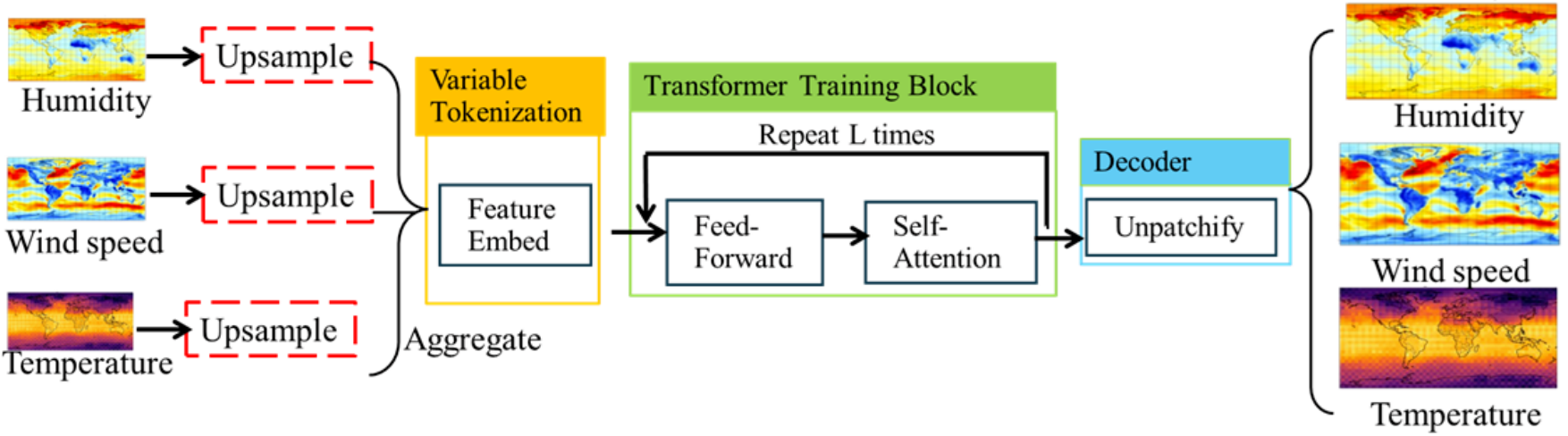}
\caption{A generalized AI architecture diagram for state-of-the-art downscaling foundation models. Note that upsampling is used for each channel prior to training blocks to reduce downscaling uncertainty.}
\vspace*{-0.4cm}
\label{fig:soa}
\end{figure*}

To overcome the limitations of task-specific models, recent efforts have introduced foundation models such as Prithvi~\cite{prithvi} and ClimateLearn~\cite{climatelearn}, which employ multi-task Vision Transformer (ViT)-based architectures to support downscaling across variables and geographic regions. While these models represent an important step forward, they remain constrained by resolution limits, computational cost, and model scalability.

A primary bottleneck is the computational complexity of downscaling at high resolution. For instance, Prithvi achieves 12 km resolution over Europe but is restricted to 50–60 km globally due to the quadratic scaling of ViT self-attention~\cite{han2023}.
ViTs divide spatial data into patches, treating each patch as a token, and self-attention computes pairwise interactions among all tokens. As resolution increases, the number of token pairs grows quadratically, resulting in quadratic growth in memory and compute demands. Unlike Natural Language Processing (NLP) models, which operate on one-dimensional text sequences and scale to over one million tokens~\cite{ulysses}, ViTs handle high-dimensional spatial inputs with complex dependencies across multiple axes, making long-sequence scaling significantly more computationally intensive. As a result, the longest ViT sequence reported to date is limited to 188k tokens~\cite{tsaris2024sequencelengthscalingvision}, directly constraining the maximum resolution, as sequence length scales proportionally with spatial resolution.

Another major challenge in downscaling is the inherent uncertainty, as mapping coarse-resolution data to fine scales is a highly ill-posed inverse problem: a single coarse input can correspond to many plausible fine-scale solutions, making the mapping non-unique and sensitive to small perturbations. This challenge is further amplified when downscaling multiple climate variables simultaneously. Unlike super-resolution tasks in computer vision~\cite{li2022srdiff, lu2022transformer}, where Red-Green-Blue channels represent the same physical quantity, climate variables such as temperature, humidity, and wind are governed by distinct yet interrelated physical processes. This heterogeneity increases the difficulty of learning consistent mappings and exacerbates uncertainty in predictions. A common mitigation strategy is to upsample coarse inputs prior to training~\cite{prithvi, climatelearn}, which can help reduce uncertainty but significantly increases sequence length and, in turn, computational cost due to ViT's quadratic complexity. Moreover, upsampling introduces artifacts that can propagate through the model, limiting its effectiveness.

Existing foundation models also face restricted model scale. For example, Prithvi~\cite{prithvi} is constrained to 1.4 billion parameters, primarily due to the computational difficulty of scaling ViTs for high-dimensional spatiotemporal data. A major advancement is the Oak Ridge Base AI foundation model for Earth System Predictability (ORBIT)~\cite{wang2024orbitoakridgebase}, which leverages hybrid sharding and orthogonal parallelisms to scale ViTs to 113 billion parameters—five times larger than previous ViTs and more than 100× larger than typical climate models. However, ORBIT is specifically designed for temporal forecasting and does not address spatial downscaling. In particular, it does not resolve the ViT long-sequence bottleneck nor mitigate the uncertainty associated with inverse downscaling problems.

\begin{figure*}[t]
\includegraphics[width=\linewidth]{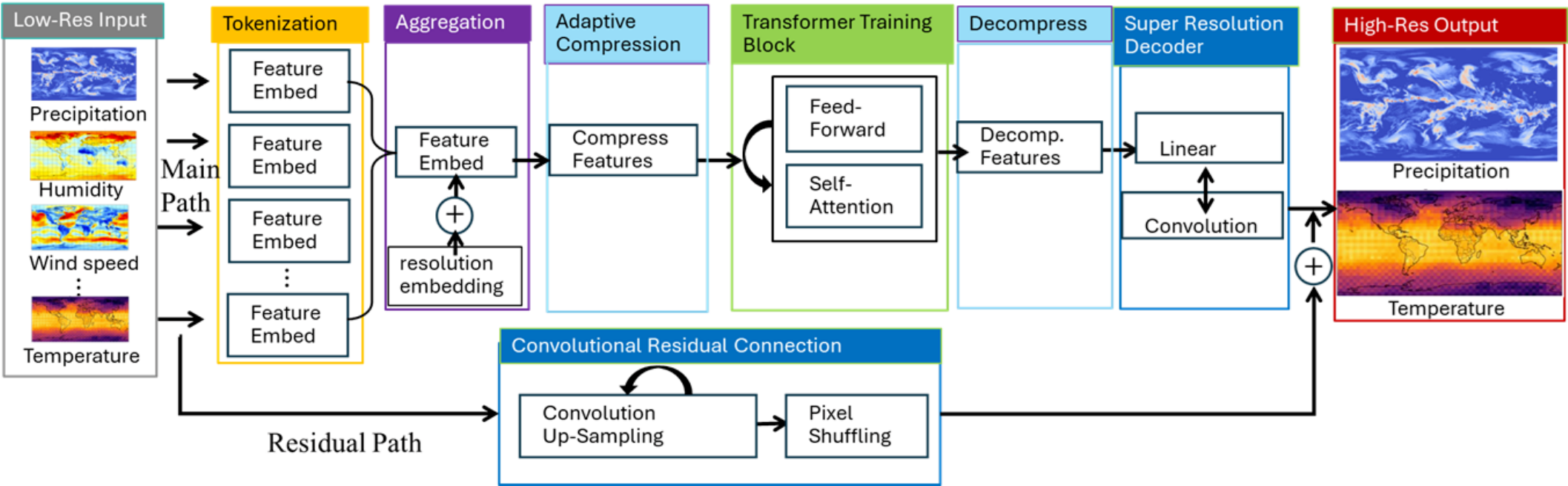}
\caption{Reslim architecture is split into main and residual paths. No upsampling is used for the main path for ViT training, leading to reduced computations. Residual path is used to condition prediction for reduced uncertainty.}
\vspace*{-0.4cm}
\label{fig:reslimvit}
\end{figure*}

To overcome these limitations, we introduce ORBIT-2, a scalable and efficient foundation model for climate downscaling.
Unlike ORBIT, which focuses on temporal weather forecasting, ORBIT-2 addresses a fundamentally different challenge: high-resolution spatial downscaling. At its core, ORBIT-2 features a novel ViT architecture, {\em Residual Slim ViT (Reslim)}, specifically designed to bypass the high computational cost associated with traditional upsampling-based approaches. Unlike existing models that upsample inputs to mitigate uncertainty—resulting in quadratic increases in memory and computation—Reslim operates directly on adaptively compressed spatial inputs, significantly reducing sequence length while preserving critical information.  It preserves accuracy and reduces uncertainty through a lightweight residual learning architecture, enabling efficient, low-overhead predictions. Additionally, both training and inference are framed as a {\em Bayesian Estimation} problem, incorporating a Markov Random Field Total Variation prior to further constrain uncertainty and improve spatial consistency.

Complementing this architecture is the {\em Tile-Wise Sequence Scaling Algorithm (TILES)} that reduces ViT’s self-attention complexity from quadratic to linear. It works by dividing images into overlapping tiles, each processed in parallel on separate Graphical Process Units (GPUs) using localized self-attention. Each tile's downscaled outputs are then seamlessly merged to the full image. This strategy enables efficient and scalable ViT-based downscaling, making ultra-high-resolution, global-scale applications computationally feasible.

Leveraging the above innovations, ORBIT-2 sets a new benchmark through four key breakthroughs:

\begin{itemize}[leftmargin=*] \item \textbf{Efficient Reslim Architecture} by operating directly on compressed inputs, achieving over 660$\times$ speedup compared to standard ViTs—without compromising accuracy.
\item \textbf{Longest ViT Sequence Length} by scaling ViT sequence lengths to unprecedented levels—up to 4.2 billion tokens for a 9.5M parameter model and 671 million tokens for a 10B model—surpassing the prior state-of-the-art of 188K tokens by several orders of magnitude~\cite{tsaris2024sequencelengthscalingvision}. This eliminates the long-standing sequence bottleneck, enabling global downscaling at resolutions as fine as 0.9 Kilometer (km).
\item \textbf{Scalable Large Model Training} by training models with up to 10 billion parameters across 65,536 GPUs, achieving 74–98\% strong scaling efficiency and sustained throughput of up to 4.1 ExaFLOPS in BF16 precision.
\item \textbf{State-of-the-Art Accuracy} achieving $R^2$ scores of 0.98 for precipitation and 0.99 for temperature at 7 km resolution over both the continental United States and global modeling, setting a new standard in high-fidelity downscaling. \end{itemize}

\section{Background \& State of The Art}
\label{sec:SOA}

Fig.~\ref{fig:soa} illustrates the generalized architecture of leading downscaling foundation models, including Prithvi~\cite{prithvi} and ClimateLearn~\cite{climatelearn}. The inputs consist of low-resolution data with multiple atmospheric physical variables, normalized and bias corrected, and each channel of the architecture reads data for a distinct variable. 
To address downscaling inverse problem uncertainty, current models upsample coarse-resolution inputs, either via interpolation~\cite{climatelearn} or convolution~\cite{prithvi}, before training.  This upsampling process is crucial, as it provides a higher-resolution baseline for ViT training, mitigating uncertainty from the inherently ill-posed nature of the multi-variable downscaling problem, thereby improving accuracy and uncertainty.
Once upsampled, multi-channel inputs are aggregated into a single-channel representation in feature space, a step that can be performed using either cross-attention mechanisms~\cite{nguyen2023climax} or shallow convolutional layers~\cite{prithvi,climatelearn}. This aggregated representation is then trained by the ViT training blocks, consisting of self-attention and feedforward sub-layers. Finally, the trained output is projected back from feature to image space for each individual physical variable.

This approach, however, introduces major challenges. Upsampling coarse-resolution input data before training increases the sequence length, which increases in proportion to the resolution increase, causing a quadratic increase in memory and computations due to ViT’s self-attention mechanism. This severely limits scalability and resolution, leaving the long-sequence bottleneck unresolved. Prithvi, for example, is limited to relatively coarse 50-60 km resolution for global downscaling. To address this, prior work proposed both AI architecture and scaling algorithm solutions.

\underline{Architecture solutions.} To mitigate this, architectures like Swin Transformer alleviate some of the computational burden by introducing a hierarchical architecture with shifted window attention~\cite{liu2021swinvit, liu2022swinvit2}. Instead of processing the entire image at once, Swin Transformer partitions the image into smaller, non-overlapping local windows, where self-attention is computed independently within each window. To capture global spatial dependencies, features learned from local windows are aggregated into global features through an architecture hierarchy.
While this reduces computing complexity, Swin Transformer has fundamental limitations and its layers of architecture hierarchy must scale proportionally with higher resolution, making it unsuitable for foundation models that needs a single model to generalize across diverse datasets with varying resolutions. 
Additionally, Swin Transformer's model size grows with the architecture hierarchy, shifting the computational bottleneck from long-sequence processing to large-model scaling. Consequently, Swin Transformer can only scale up to 147K sequence length on standard 3-channel images~\cite{liu2022swinvit2}, far below what is needed for high-resolution, multi-variable downscaling. 

Other sparse attention architectures, such as MaxViT~\cite{maxvit}, attempt to mitigate computational cost by sampling self-attention computations. While this reduces complexity, it comes at the expense of accuracy degradation when the sampling ratio is too high, and it does not address the fundamental quadratic complexity long-sequence problem.

\underline{Scaling algorithm solutions.} Besides architecture innovations, scaling algorithms, such as sequence parallelism~\cite{tsaris2024sequencelengthscalingvision, wang2023ultralongsequence, Lyngaas24}, has been proposed as an alternative strategy for scaling ViT sequence length. It distributes image patch tokens across GPUs for parallel computing, alleviating memory constraints. However, because self-attention requires each token to interact with all other tokens from every other GPU, sequence parallelism incurs substantial inter-GPU communication overhead and limits its scalability.
More critically, it does not resolve the fundamental quadratic complexity, which causes computational costs to grow rapidly with increased downscaling resolution. 
As a result, current ViT sequence parallelisms are limited to a maximum of 188K token sequence lengths~\cite{tsaris2024sequencelengthscalingvision}, which remain insufficient for high-resolution multi-variable downscaling. 

It is also important to note that other commonly used parallelisms—such as Fully Sharded Data Parallelism (FSDP)~\cite{FSDP23}, Tensor~\cite{shoeybi2020megatronlm}, pipeline~\cite{he2021pipetransformer,huang2019gpipe, kim2020torchgpipe} and hybrid sharded parallelisms~\cite{wang2024orbitoakridgebase} are all designed to scale model sizes, rather than long sequences of high-resolution and high-dimensional spatial data. Consequently, none of the existing model parallelisms overcome the long-sequence bottleneck in ViTs required for high-resolution global downscaling and there is an urgent need to develop computing efficient and massively parallel architecture and scaling algorithm.

\section{Innovation Realized}
\label{sec:innovation}

\subsection{Reslim Architecture}
Unlike existing foundation models that rely on input upsampling to establish downscaling baselines, which leads to increased sequence length and high computational cost, ORBIT-2 introduces Residual Slim ViT (Reslim), a highly efficient architecture that significantly reduces training time and memory usage without compromising accuracy. The key innovation of Reslim is its ability to operate directly on low-resolution and adaptively compressed inputs, drastically reducing sequence length and computational burden. To counteract the uncertainty typically introduced by bypassing upsampling prior to ViT training, Reslim incorporates Bayesian estimation and a residual convolutional learning path, enabling high accuracy while maintaining efficiency. Its non-hierarchical design further promotes generalization across datasets with varying spatial resolutions, making it well-suited for scalable, foundation-level Earth system modeling.

\underline{Main ViT Path.}
Fig.~\ref{fig:reslimvit} illustrates the Reslim architecture. After tokenizing each low-resolution physical variable into feature embeddings, the model proceeds along two architectural paths: the main ViT and residual paths.
Crucially, the main path eliminates input upsampling, avoiding the sequence length inflation and the quadratically increased computing cost typical of ViT architectures.

First, the main path uses a cross-attention module to aggregate multi-variable embeddings into a unified representation, effectively collapsing the variable dimension. A learnable resolution embedding encodes the desired output resolution and is added to the feature embedding, enabling resolution-aware predictions—an essential capability for modeling resolution-dependent Earth system behaviors.
Next, an optional adaptive spatial compression module, which will be explained further in the next paragraph, reduces the sizes of the embeddings before they are passed through ViT training blocks. When enabled, this module compresses spatial features; otherwise, it acts as an identity function. After processing, a decoder comprising convolutional layers and linear projections reconstructs the high-resolution output.

\underline{Adaptive Spatial Compression.}
Our objective is not only to train directly on low-resolution inputs, but also to further reduce token count and computational cost through compression. Reslim achieves this via an adaptive spatial compression technique, inspired by adaptive image patching and mesh refinement methods~\cite{adaptive_patching}. After aggregating multi-variable features (purple block in Fig.~\ref{fig:reslimvit}), the model projects the embedding back into image space and recursively partitions it into spatial quadrants using a quad-tree structure. Partitioning continues for any quadrant where the estimated feature density—computed via Canny edge detection—exceeds a predefined threshold, terminating when a minimum patch size is reached or below predefined threshold.

\begin{figure}[t]
\centering
\includegraphics[width=.9\linewidth]{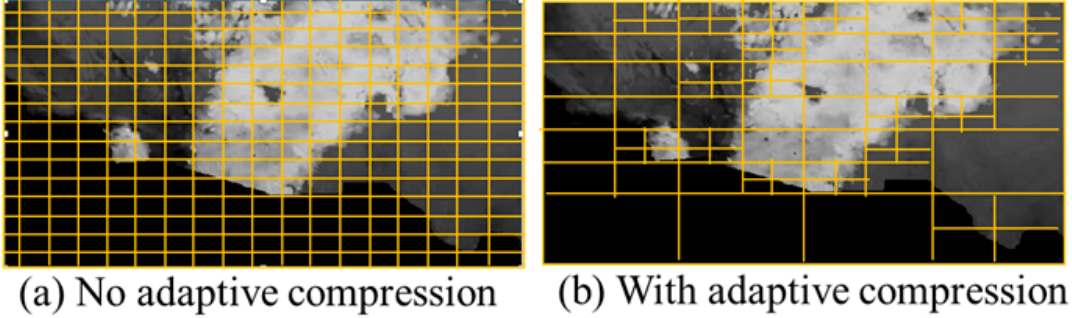}
\vspace*{-0.4cm}
\caption{Comparison with and without adaptive spatial compression. Each yellow grid is an image patch.}
\vspace*{-0.4cm}
\label{fig:adaptive_compression}
\end{figure}

This approach enables finer-grained learning in feature-rich regions through smaller patches, and coarse-grained learning to smoother regions through larger patches, where less detail is needed. Fig.~\ref{fig:adaptive_compression} illustrates an example image after variable-aggregated features are mapped back to image space. Compared to conventional uniform patching (Fig.\ref{fig:adaptive_compression}(a)), where each grid represents an image patch token, the adaptive spatial compression method (Fig.\ref{fig:adaptive_compression}(b)) reduces the number of patch tokens by 7x in this figure example, significantly decreasing sequence length and computing cost. After ViT training blocks, the decompression module reconstructs the high-resolution output from the compressed embeddings.

\underline{Residual Learning.} Reslim improves computational efficiency by removing the upsampling step from the main ViT path and training directly on low-resolution, spatially compressed inputs. This design dramatically shortens sequence lengths and reduces the quadratic computational cost typically associated with ViT training. However, bypassing input upsampling introduces uncertainty, as conventional foundation models rely on upsampled inputs to provide a coarse downscaling baseline. Reslim addresses this challenge through two complementary innovations: residual convolutional learning and a Bayesian estimation objective.

The residual convolutional path reintroduces upsampling outside the main ViT path, using lightweight convolutional layers with linear complexity. This path generates a high-resolution approximation that is added to the ViT output before loss computation. Such design yields two major benefits: (1) it avoids the expensive quadratic cost of increasing the ViT sequence length due to upsampling. The upsampling is moved to the residual path, where convolutional layers have linear complexity to input size and thereby upsampling in the residual path incurs minimal computing cost. (2) it simplifies the learning task by letting the ViT focus on predicting the residual difference between the convolutional approximation and the ground truth, rather than the full downscaling transformation. This soft constraint stabilizes training, enhances physical plausibility, and significantly reduces downscaling uncertainty. As a result, Reslim achieves high downscaling accuracy with significantly reduced computations compared to conventional ViT.

\underline{Bayesian Training Loss.} To further reduce uncertainty and improve accuracy, Reslim reformulates its training as the following Bayesian optimization problem with a Generalized Markov Random Field Total Variation prior:
\begin{equation*}
\hat{x} \gets \argmin_{\hat{x}} \| y- \hat{x} \|_{D}^2 + \sum_{k=1}^K \sum_{i=1}^N \sum_{j \in C(\hat{x_{k,i}})} b_{i,j} \| x_{k,i} - x_{k,j} \| \ ,
\label{eqn:bayesian}
\end{equation*}
where $y$ is the high-resolution ground-truth, $\hat{x}$ is the Reslim prediction, and $D$ is a latitude weighting matrix to account for the decrease in longitudinal spacing toward the poles. $K$ is the number of output variables, and $N$ is the total number of pixels per variable.
The neighborhood $C(\hat{x_{k,i}})$ contains all spatial neighbors of pixel
$\hat{x_{k,i}}$, which is the $i^\textit{th}$ pixel for the $k^\textit{th}$ variable. $b_{i,j}$ is a spatial weighting factor inversely proportional to the euclidean distance between each pixel pair in the same neighborhood.  In the above formulation, the first term, $\| y- \hat{x} \|_{D}^2 $, is the Bayesian forward data likelihood term using a latitude-weighted mean squared error. The second term is a total variation spatial prior, promoting spatial smoothness by penalizing irregularities within local neighborhoods, but also preserving edges and discontinuities. This makes it well suited for downscaling tasks for spatial coherence and structure preservation.


\subsection{TILES: Tilewise Efficient Sequence Scaling Algorithm}

\begin{figure}[t]
\includegraphics[width=.9\linewidth]{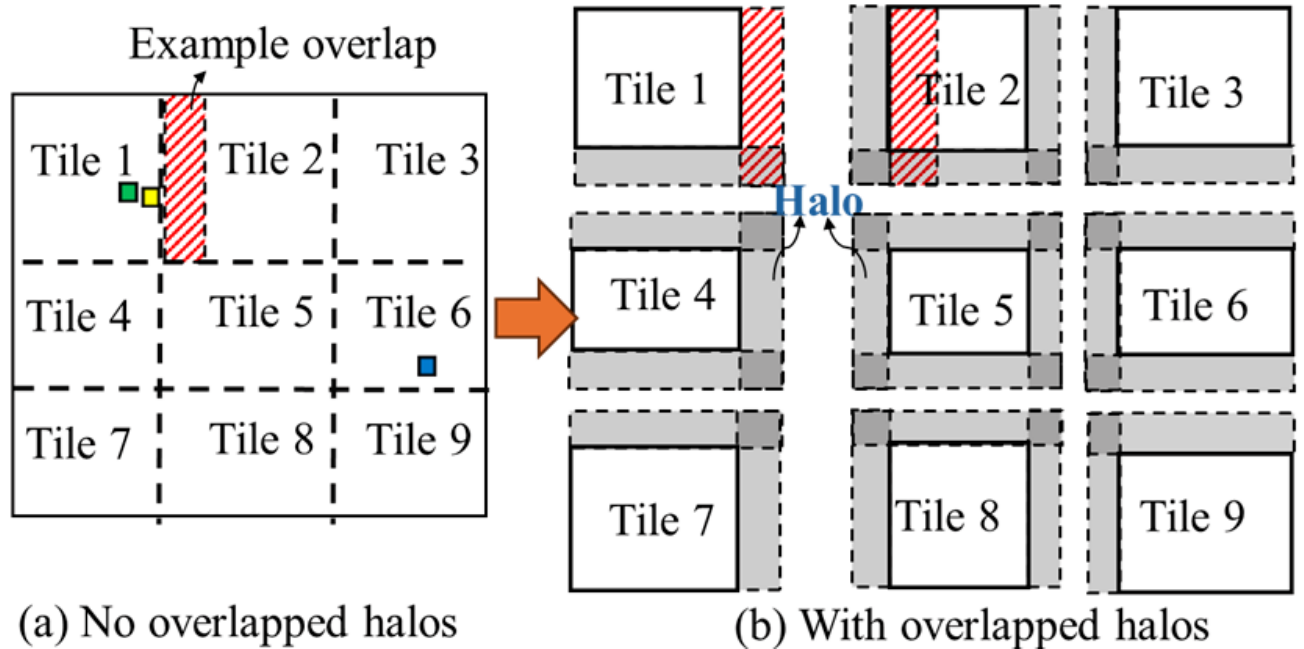}
\caption{(a) TILES algorithm partitions inputs and outputs into tiles, each processed independently on a GPU. Yellow/green pixels denote strong local correlations, while green/blue indicate weaker long-range correlations. (b) Gray halos are added around tiles to avoid border artifacts. Boundary regions (example in red) are duplicated, appearing once in a tile’s halo and once in its neighbor’s interior.}
\vspace*{-0.4cm}
\label{fig:sequence_par}
\end{figure}

While Reslim significantly reduces computation by operating on low-resolution and compressed inputs, it does not resolve the inherent quadratic complexity of self-attention. As resolution increases, this limitation becomes a bottleneck. To address this, we introduce the Tilewise Efficient Sequence Scaling Algorithm (TILES), a scalable sequence processing strategy that reduces attention complexity from quadratic to linear and enables efficient parallelization.

TILES is motivated by the spatial locality property of downscaling, where the downscaling for each high-resolution pixel is primarily influenced by spatially nearby coarse-resolution inputs. This ``point spread" effect, well studied in the remote sensing literature~\cite{HUANG2002203, WANG2020251, WANG2020112054}, shows that the correlation between pixel pairs decay rapidly when the spatial distance. Consequently, long-range pixel interactions that have weak point spread effect can be safely ignored without affecting downscaling accuracy. For example, neighboring green and yellow pixels in Fig.~\ref{fig:sequence_par}(a) have high mutual influence, while distant pairs (e.g., green and blue) contribute minimally or no influence to each other’s predictions.

Leveraging this property, TILES partitions both inputs and downscaled outputs into spatial tiles, assigning each tile to a separate GPU. Each GPU then performs downscaling separately for its assigned tile, and self-attention is restricted within each tile, preserving local context while ignoring long-range dependencies across tiles. This tilewise downscaling reduces self-attention complexity from quadratic to linear. More specifically, the computation complexity is $O(\frac{N^2}{T})$, where $N$ is the number of image patch tokens and $T$ is the number of tiles. For fixed-size tiles, $T$ increases proportionally with $N$, making the overall complexity linear.

However, strict tiling introduces border artifacts, as pixels near the border of each tile lack context from neighboring tiles. To mitigate this, TILES introduces halo padding, where each tile is extended with a fixed-width halo (shown in gray in Fig.\ref{fig:sequence_par}(b)) that overlaps adjacent tiles. For instance, the red boundary region in Fig.\ref{fig:sequence_par}(a) between Tiles 1 and 2 is duplicated, with one copy in Tile 1’s halo and another copy in Tile 2’s non-halo interior, both shaded in red in Fig.\ref{fig:sequence_par}(b). This duplication restores spatial continuity across tile boundaries, ensuring that border pixels—such as the yellow pixel—receive complete local context. Note that the halo width is tuned empirically. While in principle it can be derived directly from the data sensor’s point spread function, in practice we begin with minimal halos and increase their size iteratively until both boundary artifacts and training loss are minimized.

After each GPU independently downsamples its tile, the halo regions are discarded, and the non-padded tile outputs are stitched together to form the final high-resolution output. Since each GPU processes a different tile, leading to different gradient and model parameter update, gradients from all GPUs are averaged to maintain the model consistency across GPUs and global optimization. This inter-GPU communications, however, have minimal communication overhead as it takes place only once per data batch.

\begin{figure}[t]
\centering
\includegraphics[width=.7\linewidth]{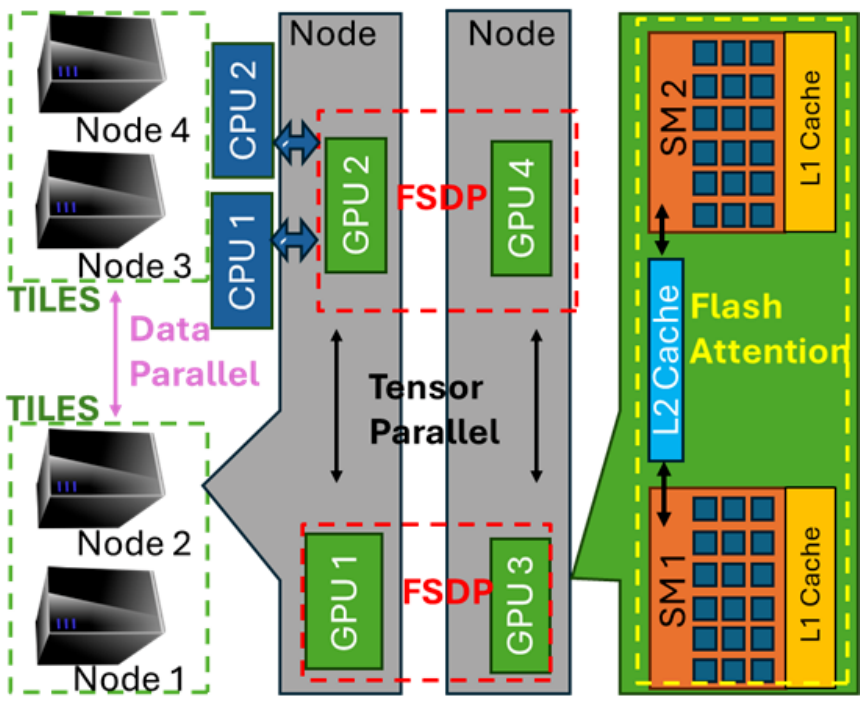}
\caption{Orthogonal levels of parallelisms mapped to supercomputer at cluster, node, and device levels.}
\vspace*{-0.4cm}
\label{fig:cluster-node-device}
\end{figure}

\begin{table*}[!h]
\centering
\small
\captionsetup{font=small}
\begin{tabular}{|l|c|c|c|c|c|c|c|c|c|}
\hline
\textbf{Dataset} & \textbf{Type} & \textbf{Interval} & \textbf{Region} & \textbf{Res. (km)} & \textbf{In/Out Vars} & \textbf{Sample Dim. (in→out)} & \textbf{\# Pairs} & \textbf{Size (GB)} \\ \hline
\multicolumn{8}{|c|}{\bf Pretraining} \\ \hline
ERA5 $\to$ ERA5 & Reanalysis & Hourly & Global & 622→156 & 23 / 3 & [32,64,23]→[128,256,3] & 367,920 & 200 \\ \hline 
ERA5 $\to$ ERA5 & Reanalysis  & Hourly & Global & 112→28  & 23 / 3 & [180,360,23]→[720,1440,3] & 367,920 & 6,328 \\ \hline 
PRISM $\to$ PRISM & Observation & Daily & U.S.   & 16→4    & 7 / 3 & [180,360,7]→[720,1440,3] & 14,235  & 189 \\ \hline 
DAYMET $\to$ DAYMET & Observation & Daily & U.S.   & 16→4    & 7 / 3 & [180,360,7]→[720,1440,3] & 14,946  & 200 \\ \hline
\multicolumn{8}{|c|}{\bf U.S. Regional Fine-Tuning for Tmin, Tmax and Precipitation} \\ \hline
\begin{tabular}{@{}c@{}} [ERA5, DAYMET] \\ $\to$ DAYMET \end{tabular} & \begin{tabular}{@{}c@{}} Reanalysis \\ \& Observation \end{tabular} & Daily & U.S. & 28→7 & 23 / 3 & [180,360,23]→[720,1440,3] & 14,946 & 254 \\ \hline
\multicolumn{8}{|c|}{\bf Global Precipitation Fine-Tuning} \\ \hline
\begin{tabular}{@{}c@{}} [ERA5, IMERG] \\ $\to$ IMERG \end{tabular}
& \begin{tabular}{@{}c@{}} Reanalysis \\ \& Observation \end{tabular} & Daily & Global & 28→7 & 23 / 1 & [720,1440,23]→[2880,5760,3] & 8,401 & 745 \\ \hline
\end{tabular}
\caption{Datasets used for pretraining, fine-tuning, and inference. Each entry specifies the downscaling resolution, input/output variable counts, sample dimensions, number of training samples, and total storage size.}
\vspace*{-0.4cm}
\label{tab:datasets}
\end{table*}

\subsection{Orthogonal Parallelisms}
TILES efficiently scales sequence length, enabling high-resolution downscaling. However, it does not address model size scaling. To support both large foundation models and high-resolution downscaling, TILES must be integrated with complementary model-parallel strategies. Since TILES and model parallelism target orthogonal goals, with TILES for sequence length and model parallelisms for model size, they can be combined seamlessly. This results in a unified framework with four distinct parallelism strategies:

\begin{itemize}[leftmargin=*]
      \item \textbf{TILES sequence parallelism}: Distributes long sequence lengths for ViTs for tilewise approximation as discussed before. Requires least communication overhead.
  \item \textbf{Fully Sharded Data Parallelism} (FSDP)~\cite{FSDP23}: shards both data and model parameters across GPUs, but requires temporarily gathering the full model during forward and backward passes. Requires moderate communication overhead.
  \item \textbf{Tensor Model Parallelism}~\cite{shoeybi2020megatronlm}: Only shards model parameters and keep parameters sharded throughout training. Requires most communication overhead.
  \item \textbf{Distributed Data Parallelism} (DDP)~\cite{li2020pytorch}: distributes only the training data without sharding model parameters. Requires least communication overhead.
\end{itemize}

Fig.~\ref{fig:cluster-node-device} shows how these orthogonal parallelisms map to a supercomputer hardware. Two adjacent nodes form a TILES sequence parallel group (green dashed boxes), responsible for scaling sequence lengths. Multiple sequence parallel groups form a DDP group, distributing data batches across the system.

Within each sequence parallel group, GPUs participate in both tensor model and FSDP parallelisms for model scaling. Tensor model parallelism operates within a node, leveraging its low-latency interconnect to mitigate communication overhead for hidden dimension partitioning. FSDP (red dashed boxes) spans GPUs across neighboring nodes within the same TILES group, enabling parameter and data sharding.

Once all parallelisms are established, each GPU receives a subset of the model and data that can be optimized with Flash Attention~\cite{dao2023flashattention2} to reduce memory use and cache misses, as detailed in the next subsection. Within each GPU, streaming multiprocessors (SMs) are mapped to Flash Attention cache blocks, executing vector operations in parallel within each block. Meanwhile, CPUs asynchronously load data and construct quad-trees to track the spatial layout of adaptively compressed patches described in Fig.~\ref{fig:adaptive_compression}.

Note that this multi-level strategy aligns the parallelism hierarchy with the hardware architecture to optimize performance. Neither DDP nor TILES sequence parallelisms  requires frequent communication, and are therefore mapped to cluster nodes with slower network communication. Tensor and FSDP model parallelisms requires more frequent communication, and are therefore mapped to GPUs within the same node and across neighboring nodes to utilize their faster in-node and neighboring-node network communications. The Flash Attention require the most frequent communication and are therefore mapped to SMs within the same GPU, which has the fastest network through shared L2 cache.

\subsection{Optimizations}
To further boost performance, we applied these optimizations:

\underline{Hybrid-OP Parallelism}.
We adopt the Hybrid-OP optimization technique from ORBIT~\cite{wang2024orbitoakridgebase}, which leverages the mathematical structure of matrix chain multiplication to shard model parameters in alternating row and column dimensions. This optimization combines tensor model parallelism with FSDP, achieving superior scalability with reduced communication overhead and frequency compared to without Hybrid-OP.

\underline{Flash Attention}.
To accelerate self-attention computation, we use Flash Attention~\cite{dao2023flashattention2}, which applies a cache-blocking technique to minimize memory access to GPU global memory. By maximizing data reuse from high-bandwidth on-chip cache, Flash Attention significantly improves compute throughput through higher cache hit rates and faster memory access.

\underline{Mixed Precision and Layer Wrapping}.
We further utilize BFLOAT16 mixed-precision to speed up training while reducing memory usage. To address numerical instability—where gradients with extreme magnitudes may underflow or overflow in BFLOAT16—we apply PyTorch’s dynamic gradient scaling~\cite{pytorch-gradscaling}. This technique automatically rescales gradients into a representable range and reverses the scaling during parameter updates, ensuring numerical stability.

To further reduce communication cost, we apply FSDP in a layer-wise fashion~\cite{wang2024orbitoakridgebase}. Instead of sharding all model layers in a single instance, parameters are sharded one layer at a time. This reduces synchronization overhead and memory use.

\begin{table*}[t]
\centering
\small
\captionsetup{font=small}
\begin{tabular}{|l|c|c|c|c|c|c|c|c|c|}
\hline
\textbf{Arch} & \textbf{Model Size} & \textbf{Resolution (km)} & \textbf{Seq. Length} & \textbf{Compression} & \textbf{Tiles} & \textbf{Time/sample (s)} & \textbf{Speedup} & \textbf{PSNR} & \textbf{SSIM} \\
\hline
\multicolumn{10}{c}{\bf (a) Reslim Architecture Speedup Comparison with ViT} \\ \hline
ViT     & 9.5M  & 622$\rightarrow$156  & 24,576    & 1×   & 1   & 7.3e-4 & 1   & 35.0 & 0.94 \\ \hline
Reslim  & 9.5M  & 622$\rightarrow$156  & 24,576    & 1×   & 1   & 1.1e-6 & 660   & 36.7 & 0.96 \\ \hline
ViT     & 9.5M  & 112$\rightarrow$28   & 777,660   & 1x  & 1   & OOM & NA       & NA   & NA   \\ \hline
Reslim  & 9.5M  & 112$\rightarrow$28   & 777,660   & 1×   & 1   & 1.2e-3 & NA     & 37.6 & 0.96 \\ \hline
\multicolumn{10}{c}{\bf (b) Adaptive Compression \& Tiling Speedup Comparison with Reslim Baseline} \\ \hline
Reslim  & 9.5M  & 112$\rightarrow$28   & 777,660   & 8×   & 1   & 3.6e-4 & 3.3  & 37.7 & 0.96 \\ \hline
Reslim  & 9.5M  & 112$\rightarrow$28   & 777,660   & 16×  & 1   & 1.8e-4 & 6.6   & 37.8 & 0.96 \\ \hline
Reslim  & 9.5M  & 112$\rightarrow$28  & 777,660   & 32×  & 1   & 1.7e-4 & 7.1  & 37.9 & 0.96 \\ \hline
Reslim  & 9.5M  & 112$\rightarrow$28  & 777,660   & 1×   & 4   & 8.0e-4 & 1.5  & 37.7 & 0.96 \\ \hline
Reslim  & 9.5M  & 112$\rightarrow$28   & 777,660   & 1×   & 16  & 6.3e-4 & 1.9  & 37.7 & 0.96 \\ \hline
Reslim  & 9.5M  & 112$\rightarrow$28   & 777,660   & 1×   & 36  & 7.4e-4 & 1.6  & 37.7 & 0.96 \\
\hline
\end{tabular}
\caption{(a) Computing performance comparison between vanilla conventional ViT and Reslim at 128 GPUs. For 622$\rightarrow$156 km downscaling, Reslim achieves a 660× speedup over conventional ViT while maintaining accuracy. (b) Illustrates Reslim’s performance gains at varying adaptive compression rates and tile counts, relative to a Reslim baseline without compression or tiling.}
\vspace*{-0.4cm}
\label{tab:ablation-study}
\end{table*}

\begin{figure*}[t]
\centering
\includegraphics[width=.9\linewidth]{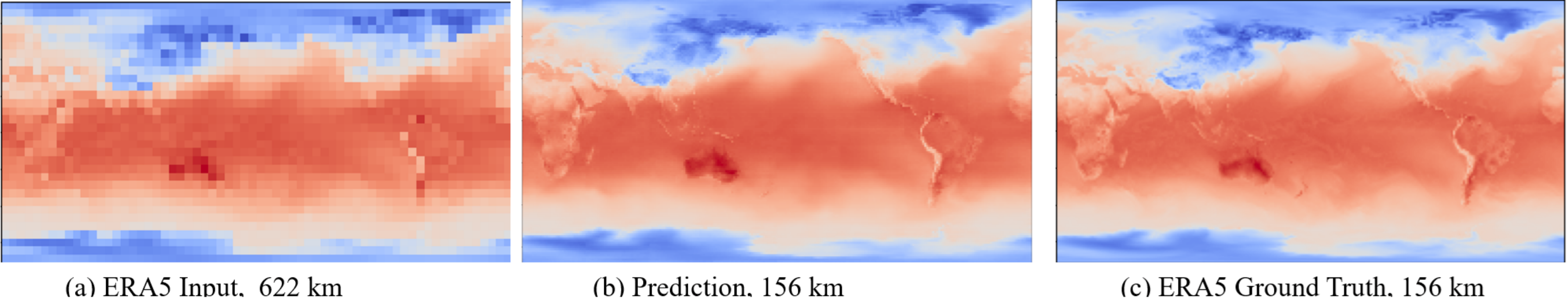}
\vspace*{-0.2cm}
\caption{Illustration of ERA5 surface temperature downscaling: (a) input at 622 km resolution, (b) ORBIT-2's downscaled prediction at 156 km, and (c) ground-truth at 156 km.  
}
\vspace*{-0.4cm}
\label{fig:downscale_example}
\end{figure*}

\section{How Performance Was Measured}
\label{sec:experiment setup}
\noindent\ul{\textbf{Model Configuration.}}
All experiments in Secs.~\ref{sec:computing-expriments} and~\ref{sec:science-expriments}  use four model configurations: 9.5M (256-dim embedding, 6 layers, 4 heads), 126M (1024-dim, 8 layers, 16 heads), 1B (3072-dim, 8 layers, 24 heads), and 10B (8192-dim, 11 layers, 32 heads) parameters.

\noindent\ul{\textbf{System Details.}}
All experiments were conducted on the Frontier supercomputer from Oak Ridge National Lab. Each node consists of one 64-core AMD EPYC CPU and 8 GPUs (64 GB memory each), organized into 4 MI250X cards with two GPUs per card. GPUs on the same card communicate via Infinity Fabric CPU-GPU, while all four MI250X cards are connected via 50 GB/s GPU-GPU Infinity Fabric. Nodes are interconnected using 100 GB/s Slingshot-11. Software stack includes PyTorch v2.7, ROCm v6.3.1, and libfabric v1.22.

The only exception is Table~\ref{tab:frontier_alps_perf}, which compares sustained throughput on Frontier and Alps supercomputer. Alps is built on NVIDIA GH200 Grace Hopper Superchips. Each node features 4 Hopper GPUs (96 GB memory each), with every GPU paired to a 72-core Grace CPU via a 900 GB/s NVLink-C2C interface, enabling unified CPU–GPU memory. Nodes are interconnected using NVIDIA Quantum-2 InfiniBand.

\noindent\ul{\textbf{Datasets.}}
Table~\ref{tab:datasets} summarizes the datasets used for pretraining and fine-tuning. ORBIT-2 can perform downscaling at arbitrary resolution but as a use case demonstration, our model is trained on paired input$\,\to\,$output datasets for 4x spatial refinement.

For global pretraining, we use the ERA5 reanalysis dataset (1980-2020)~\cite{era5}, which integrates historical observations with numerical simulations. Two ERA5 resolution pairs are used: 622 km$\,\to\,$156 km and 112 km$\,\to\,$28 km. The data are split into 38 years for training, 2 years for validation, and 1 year for testing.  23 input variables are used, including 5 static fields, 12 atmospheric variables (humidity, wind speed, and temperature at 200, 500, and 850 hPa), and 6 surface variables.
For United States (US)-focused pretraining, we utilize the PRISM and DAYMET observation datasets (from 1980 to 2022)~\cite{daly2013prism, thornton2014daymet}, which derived from ground-based weather stations. We perform 4x downscaling from 16 km to 4 km. 

For continental U.S. fine-tuning, inputs include both ERA5 and DAYMET at 28  km, with 7 km DAYMET as output ground truth. The fine-tuning dataset is split into training, validation, and testing in the same way as pretraining. The prediction output variables are daily minimal temperature (Tmin), maximal temperature (Tmax), and total precipitation.
For global fine-tuning, we target precipitation only and inputs include both ERA5 and IMERG precipitation data at 28 km, downscaled to 7 km resolution. IMERG~\cite{huffman2020integrated} is a satellite-based precipitation observation product from NASA. Model predictions are daily aggregated and compared to IMERG globally.

\begin{table*}[t]
\centering
\small
\captionsetup{font=small}
\begin{tabular}{|l|c|c|c|c|c|c|c|}
\hline
\textbf{Architecture} & \textbf{Model Size} & \textbf{Compression} & \textbf{Tiles} & \textbf{GPUs} & \textbf{Max Seq. Length} & \textbf{Output Size} & \textbf{Global Resolution (km)} \\
\hline
ViT      & 9.5M  & 1×  & 1   & 8    & 25K      & [128, 256, 18]        & 156   \\ \hline
ViT      & 10B   & 1×  & 1   & 8    & OOM      & —                     & —     \\ \hline
Reslim   & 9.5M  & 1×  & 1   & 8    & 298M     & [5760, 11520, 18]     & 3.5   \\ \hline
Reslim   & 9.5M  & 1×  & 1   & 32   & 466M     & [7200, 14400, 18]     & 2.7   \\ \hline
Reslim   & 9.5M  & 4×  & 16  & 8    & 1.1B     & [11520, 23040, 18]    & 1.7   \\ \hline
Reslim   & 9.5M  & 4×  & 16  & 128  & 4.2B     & [21600, 43200, 18]    & 0.9   \\ \hline
Reslim   & 10B   & 1×  & 1   & 8    & 18M    & [1440, 2880, 18]      & 14    \\ \hline
Reslim   & 10B   & 4×  & 16  & 8    & 74M      & [2880, 5760, 18]      & 6.9   \\ \hline
Reslim   & 10B   & 4×  & 16  & 512  & 671M     & [8640, 17280, 18]     & 2.3   \\ \hline
\end{tabular}
\caption{Maximum sequence length scaling across architectures, model sizes, compression, tiles and GPU count.}
\vspace*{-0.4cm}
\label{tab:max_seq_len}
\end{table*}

\begin{figure*}
\centering
\includegraphics[width=.95\linewidth]{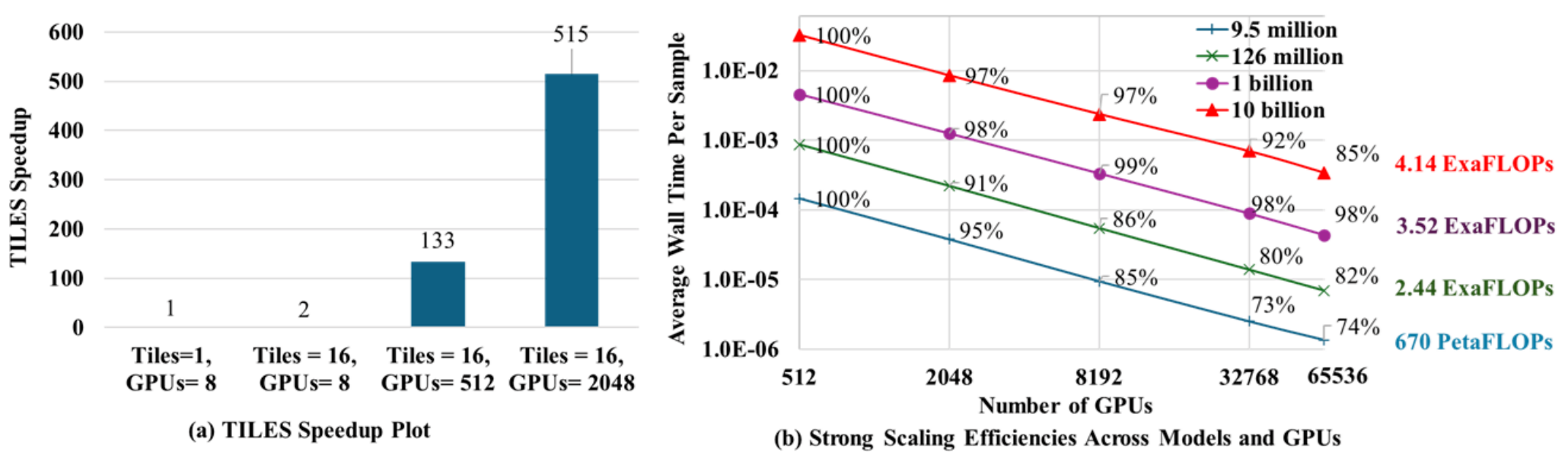}
\caption{(a) TILES sequence scaling algorithm speedup across GPUs, compared to an 8-GPU baseline that does not utilize tiling. (b) Strong scaling efficiencies up to 8192 nodes (65,536 GPUs) for various model sizes, maintaining a strong scaling efficiencies of 74-98\% at 8192 nodes.}
\vspace*{-0.4cm}
\label{fig:speedup-strong_scaling}
\end{figure*}

\noindent\ul{\textbf{Performance Metrics.}}
The total number of floating point operations (FLOPs) of the systems was collected via the Microsoft Deepspeed Profiler~\cite{deepspeed} and we only gathered the FLOPs on GPUs. Only the mixed-precision BFLOAT16 results were reported. We measured following performance:

\begin{itemize}[leftmargin=*]
\item {\textit{Time-to-solutions}}. Defined as the average wall time to downscale each sample during training. Equivalent to epoch runtime divided by the total sample count. We reported numbers for both 622$\,\to\,$156 and 112$\,\to\,$28 km resolutions.
\item {\textit{Strong scaling efficiency}}. Measured speedup per epoch relative to GPU count, with the runtime at 512 GPUs (64 nodes) as the 100\% baseline.
\item {\textit{Sustained throughput}}. It is defined as the total FLOPs to downscale each data point, divided by the average wall clock time in seconds. The performance includes the whole application with IO. Reported in ExaFLOPS.
\item {\textit{Accuracy}}. We use both scientific and image-based metrics for downscaling accuracy against observations: Coefficient of determination ($R^2$), Root-Mean-Square-Error (RMSE), RMSE for Quantiles, Structural Similarity Index (SSIM), and Peak-Signal-Noise-Ratio (PSNR). Higher $R^2$, SSIM, and PSNR scores represent higher fidelity downscaling, while lower RMSE represents higher fidelity downscaling.
\end{itemize}

\noindent\ul{\textbf{Dataset, Code and Model Checkpoint Release.}} Datasets, source code, and model checkpoints from this work will be made publicly available through the official ORNL GitHub repository. The corresponding links will be added to the arXiv version of this paper once the release process is complete.

\section{Computing Performance Results}
\label{sec:computing-expriments}

\noindent\ul{\textbf{Speedup Ablation Studies.}}
Table~\ref{tab:ablation-study}(a) presents an ablation study comparing the training performance of vanilla conventional ViT and the proposed Reslim architectures, using models with 9.5M parameters and at 128 GPUs. Two ERA5$\rightarrow$ERA5 downscaling tasks were used: 622$\rightarrow$156 km and 112$\rightarrow$28 km (details in Table~\ref{tab:datasets}). Each hourly output sample is tokenized into image patches. For the 622$\rightarrow$156 km task, outputs of shape [128, 256, 3] and 2$\times$2 patch size yield sequence length of 24,576; for 112$\rightarrow$28 km, output size [720, 1440, 3] produces 777,660 tokens. No adaptive spatial compression or tiling was applied in this comparison. For visualization, Fig.~\ref{fig:downscale_example} illustrates an example ERA5 hourly sample for surface temperature: (a) ERA5 input at 622 km, (b) downscaled prediction at 156 km using Reslim architecture, and (c) ERA5 ground truth at 156 km.

\begin{table*}[t]
\centering
\small
\captionsetup{font=small}
\begin{minipage}[t]{0.48\textwidth}
\centering
\begin{tabular}{l|cccc}
\hline
Model Size & 9.6M & 126M & 1B & 10B \\
\hline
Frontier (MI250x) & 23 & 76 & 122 & 166 \\
Alps (GH200)      & 72 & 212 & 251  & 368 \\
\hline
\end{tabular}
\caption{Sustained computing throughput (PFLOPs) comparison of Frontier and Alps on 2048 GPUs.}
\vspace*{-0.4cm}
\label{tab:frontier_alps_perf}
\end{minipage}
\hfill
\begin{minipage}[t]{0.48\textwidth}
\centering
\begin{tabular}{l|cccc}
\hline
Model Size & 9.5M & 126M & 1B & 10B \\
\hline
Inference Time (s) & 4.5e-3 & 2.6e-2 & 7.2e-2 & 5.5e-1 \\
\hline
\end{tabular}
\caption{Model inference speed on a single node (8 GPUs).}
\vspace*{-0.4cm}
\label{tab:inference_speed_single_node}
\end{minipage}
\end{table*}

The seventh column of Table~\ref{tab:ablation-study}(a) reports the average time to downscale each hourly sample. The eighth column shows the speedup from Reslim relative to the ViT baseline. Notably, the Reslim architecture avoids expensive upsampling operations by operating directly on low-resolution inputs, resulting in significant computational savings. For the smaller 622$\rightarrow$156 km task, Reslim achieves a 660× speedup over ViT at the same number of GPUs while maintaining comparable accuracy, as measured by PSNR and SSIM. This demonstrates the effectiveness of Reslim’s residual learning design and Bayesian training loss in maintaining predictive accuracy while reducing computational cost.
For the larger 112$\rightarrow$28 km resolution task, the ViT model fails due to out-of-memory (OOM) errors. Consequently, a direct speedup comparison is not available, although Reslim completes the task efficiently and maintains high accuracy.

Table~\ref{tab:ablation-study}(b) explores further speedup gains from adaptive spatial compression and sequence tiling, compared to the Reslim baseline (Table~\ref{tab:ablation-study}(a), row 4), all using 128 GPUs. Adaptive compression with a 32× sequence length reduction yields up to a 7.1× speedup with no loss in PSNR or SSIM. Further compression yields diminishing returns due to quad-tree overhead.
Tiling provides up to 1.9× with 16 tiles per sample. Further tiling introduces excessive halo padding overhead and degrades computing performance. Accuracy remains the same across all settings.

\noindent\ul{\textbf{Maximal Sequence Length Scaling.}}
Table~\ref{tab:max_seq_len} presents sequence length and resolution scaling performance of various model architectures and strategies, demonstrating how the combination of spatial compression, tiling, and the Reslim architecture enables extreme sequence lengths. We achieve sequence lengths of up to 4.2 billion tokens (global downscaling resolution of 0.9 km) for a 9.5M parameter model and up to 671 million tokens (global resolution 2.3 km) for a 10B parameter model. These results surpass the state-of-the-art in sequence scaling by more than 22,000×, compared to state-of-the-art sequence parallelism of 188K tokens~\cite{tsaris2024sequencelengthscalingvision}, and the Swin Transformer at 147K tokens~\cite{liu2022swinvit2}.

All experiments utilize 23 input variables (12 atmospheric, 6 surface, and 5 static) and produce 18 output variables (excluding static inputs). Using a conventional ViT with 9.5M parameters, the maximum sequence length is limited to 25K tokens (coarse 156 km global resolution) when using 8 GPUs. Scaling this ViT model to 10B parameters results in an out-of-memory (OOM) error, making global downscaling infeasible.

\begin{table*}[t]
\centering
\small
\captionsetup{font=small}
\begin{tabular}{|c|c|c|c|c|c|c|c|c|}
\hline
\textbf{Fine-Tuning Task} & \textbf{Model Size} & \textbf{R\textsuperscript{2}} & \textbf{RMSE} & \textbf{RMSE $\sigma_1 >68\%$} & \textbf{RMSE $\sigma_2 >95\%$} & \textbf{RMSE $\sigma_3 >99.7\%$} & \textbf{SSIM} & \textbf{PSNR} \\
\hline
\multirow{2}{*}{U.S. Tmin (Kelvin)} & 9.5M  & 0.991 & 3.812 & 4.652 & 9.704 & 15.497 & 0.958 & 29.02 \\
& 126M  & 0.999 & 0.505 & 0.630 & 1.025 & 1.491 & 0.987 & 45.96 \\
\hline
\multirow{2}{*}{U.S. Precipitation (mm/day)} & 9.5M  & 0.975 & 0.146 & 0.166 & 0.344 & 0.449 & 0.931 & 29.03 \\
& 126M  & 0.979 & 0.135 & 0.154 & 0.296 & 0.365 & 0.932 & 30.20 \\
\hline
\multirow{2}{*}{Global Precipitation (mm/day)} & 9.5M  & 0.986 & 0.098 & 0.120 & 0.136 & 0.133 & 0.923 & 35.09 \\
& 126M  & 0.986 & 0.099 & 0.122 & 0.137 & 0.138 & 0.913 & 35.04 \\
\hline
\end{tabular}
\caption{Comparison of downscaling accuracy for fine-tuning tasks for U.S. temperature (Tmin), U.S. precipitation, and global precipitation using models with 9.5M and 126M parameters. The table includes $R^2$, RMSE, RMSE at different percentiles ($\sigma_1$, $\sigma_2$, $\sigma_3$), SSIM, and PSNR.}
\vspace*{-0.4cm}
\label{tab:merged_finetuning}
\end{table*}

In contrast, Reslim demonstrates significantly better scaling. With just 8 GPUs, a 9.5M parameter Reslim model scales to 298M tokens at a 3.5 km global resolution. This corresponds to an output tensor of shape [5760, 4520, 18], assuming a 2×2 image patch size. Increasing the number of GPUs to 32, we achieve 466M tokens at 2.7 km resolution.

When combining Reslim with both spatial tiling (16 tiles per sample) and adaptive spatial compression (4×) techniques, substantial improvements are obtained. With these methods, the model achieves 1.1B tokens on only 8 GPUs, corresponding to downscaled output of size [11520, 23040, 18]. This result is made possible through several key compression techniques:
\begin{itemize}[leftmargin=*]
\item Channel aggregation in Reslim (see Fig.~\ref{fig:reslimvit}) reduces the sequence length by 18× by aggregating channels.
\item Spatial tiling divides the sample into 16 tiles, reducing the sequence length per GPU by 16×.
\item Adaptive spatial compression reduces sequence by 4×.
\item  Reslim processes directly on low-resolution input, reducing the effective sequence length by 60×.
\end{itemize}
Combining all four, the effective per-GPU sequence length becomes only 17,280 tokens, despite the global output representing 1.1 billion tokens.
Finally, by scaling to 128 GPUs, we achieve our largest configuration: 4.2 billion tokens at 0.9 km resolution.

For the 10B parameter model, Reslim still scales efficiently. Without compression or tiling, it reaches 18 million tokens. With 4× compression, 16 tiles, and 512 GPUs, the model handles 671 million tokens at 2.3 km resolution.


\noindent\ul{\textbf{TILES Sequence Scaling Speedup.}}
Fig.~\ref{fig:speedup-strong_scaling}(a) demonstrates the scalability of the TILES algorithm. With 16 tiles per sample, TILES achieves a 1.9× speedup over the non-tiling baseline, both at 8 GPUs, using a 9.5M parameter model on the ERA5$\rightarrow$ERA5 112$\rightarrow$28 km downscaling task. As GPU count increases, speedup scales nearly linearly, reaching 515× at 2048 GPUs relative to the 8-GPU baseline without tiling. This highlights the scalability and minimal overhead of the TILES approach for distributed training.

\noindent\ul{\textbf{Strong Scaling Efficiencies \& Sustained Throughput.}}
Fig.~\ref{fig:speedup-strong_scaling}(b) presents strong scaling performance across model sizes (9.5M to 10B parameters) using the same dataset as in Fig.~\ref{fig:speedup-strong_scaling}(a). Experiments were conducted at scales of 64, 256, 1024, 4096 and 8192 nodes with 8 GPUs for each node at Frontier supercomputer, employing all four forms of orthogonal parallelism (see Sec.~\ref{sec:innovation}). Each point in the figure reports the average runtime in second per hourly sample with a data label for corresponding strong scaling efficiency, relative to each model's baseline performance at 64 nodes (512 GPUs).

All model sizes maintain high strong scaling efficiencies between 74–98\%. The smallest model (9.5M) underutilizes hardware at large scales due to insufficient computing with small model, with 1.3e-6 seconds per sample and a sustained computing throughput at 670 PetaFLOPS at 8192 nodes (65,536 GPUs). In contrast, larger models saturate compute resources: the 126M, 1B, and 10B models reach sustained throughputs of 2.4, 3.5, and 4.1 ExaFLOPS, respectively, at 8192 nodes. These results demonstrate the strong scalability of Reslim and the effectiveness of orthogonal parallelism for exascale climate downscaling.

\noindent\ul{\textbf{Cross Platform Throughput Comparison.}}
Table~\ref{tab:frontier_alps_perf} compares sustained computing throughput on Frontier (AMD MI250x) and Alps (NVIDIA GH200) supercomputers, using 2,048 GPUs across varying model sizes.
The training dataset is the same as Fig.~\ref{fig:speedup-strong_scaling}, downscaling ERA5 from 112 km to 28 km resolution.
On Frontier, throughput scales from 23 PFLOPs for a 9.6M-parameter model to 166 PFLOPs for a 10B-parameter model. On Alps, throughput is higher, reaching 72 PFLOPs for the 9.6M-parameter model and 368 PFLOPs for the 10B-parameter model.

\noindent\ul{\textbf{Inference Speed.}}
All the results above focus on computing performance for training, but a key advantage of AI foundation models is their efficiency at inference. Once trained, they can be deployed on edge devices with limited resources and deliver near real-time predictions.  
We evaluate inference performance on a single Frontier node with 8 GPUs (Table~\ref{tab:inference_speed_single_node}), using the same ERA5$\rightarrow$ERA5 dataset as in Fig.~\ref{fig:speedup-strong_scaling} and Table~\ref{tab:frontier_alps_perf}.  
For the 9.5M-parameter model, downscaling each global sample requires only 4 millisecond. For the 10B-parameter model, it takes 0.55 second.  
In contrast, the non-AI numerical approaches require days or weeks of computation on a large supercomputer~\cite{Fuhrer18, Taylor23}.  
This highlights the unmatched prediction speed of AI, enabling deployment in resource-limited environments with near real-time performance.

\section{Science Performance Results}
\label{sec:science-expriments}
\noindent\ul{\textbf{U.S. Regional Fine-Tuning.}}
Following pretraining on the datasets in Table~\ref{tab:datasets}, we fine-tune ORBIT-2 models with 9.5M and 126M parameters on two tasks for science demonstration: (1) US-specific downscaling of ERA5 and DAYMET data (1980 to 2022) from 28 km to 7 km, evaluated against 7 km DAYMET observations for both daily total precipitation and Tmin; and (2) global precipitation downscaling of ERA5 and IMERG (1998 to 2022) from 28 km to 7 km, evaluated against 7 km IMERG observations. These tasks evaluate model capability in downscaling at both regional and global scale.

Table~\ref{tab:merged_finetuning}'s second row summarizes U.S. regional downscaling results for Tmin at the unseen testing years. Both 9.5M and 126M models accurately reconstruct high-resolution temperature at 7 km resolution and are capable of capturing extremes, with the larger 126M model consistently outperforming the smaller 9.5M model across all metrics with lower RMSE, and higher $R^2$, SSIM and PSNR. Notably, the 126M model achieves an $R^2$ of 0.999 and SSIM of 0.987, establishing a new benchmark for temperature downscaling at 7 km resolution.

\begin{figure}
\centering
\includegraphics[width=.7\linewidth]{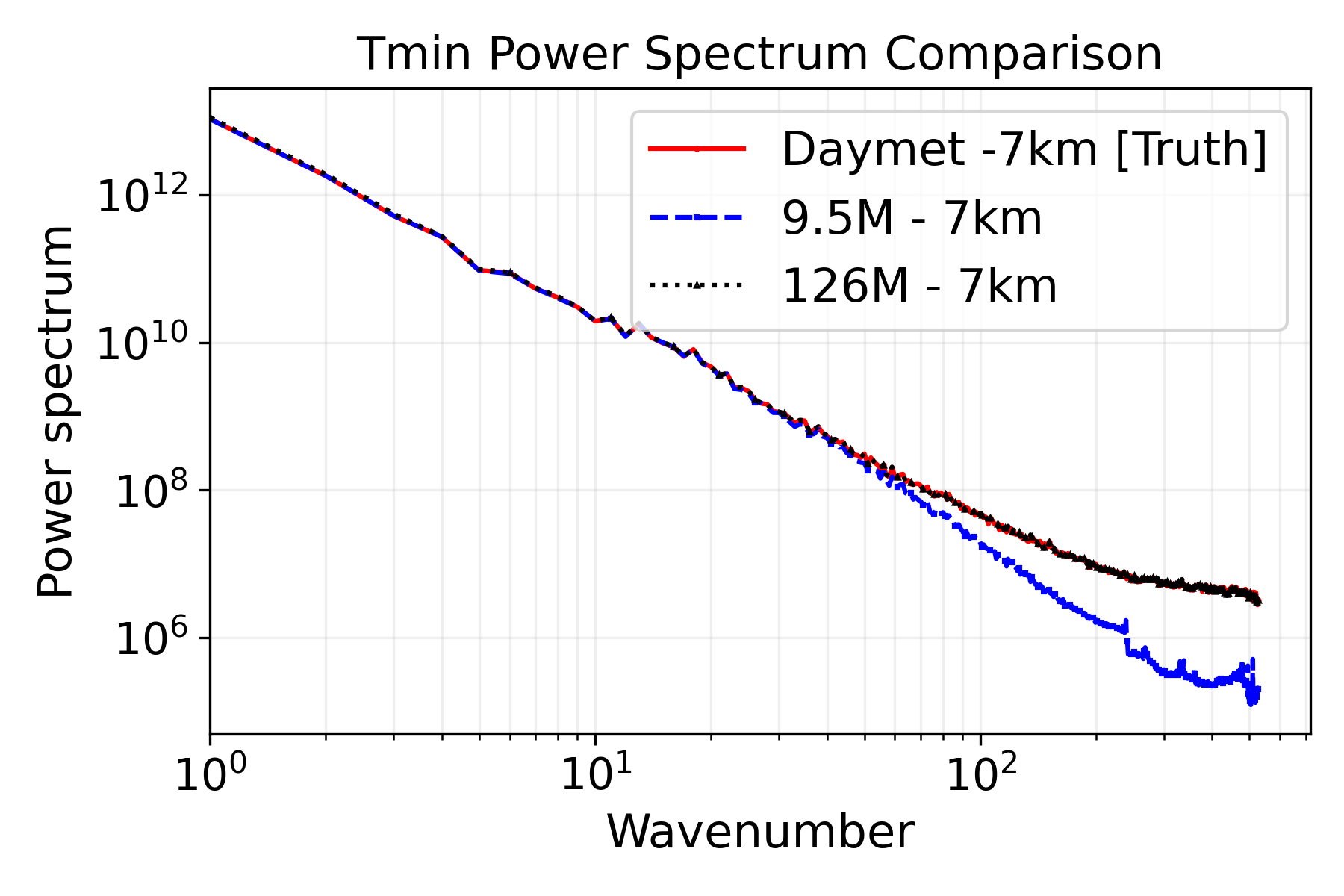}
\vspace*{-0.2cm}
\caption{Power spectra of downscaled minimum temperature using 9.5M and 126M parameter models, showing improved high-frequency fidelity with increased model capacity.}
\vspace*{-0.4cm}
\label{fig:power_spec_prcp}
\end{figure}

Fig.~\ref{fig:power_spec_prcp} provides corresponding spectral analysis for Tmin downscaling results in Table~\ref{tab:merged_finetuning}'s row two, by comparing the spatial power spectra of the two model sizes. The 126M model accurately captures high-frequency content, closely matching the high frequency spectral characteristics of the DAYMET observation ground truth. In contrast, the 9.5M model deviates from the ground truth at high frequencies. This demonstrates the larger model’s ability to resolve fine-scale spatial variability, emphasizing the value of increased model capacity for high-fidelity climate downscaling.

Table~\ref{tab:merged_finetuning}'s third row  presents downscaling results for daily total precipitation—one of the most challenging variables due to its high spatial variability and localized extremes. ORBIT-2 demonstrates strong performance, closely matching observations when downscaling to 7 km resolution. The 126M model consistently outperforms the 9.5M model across all evaluation metrics, achieving an $R^2$ of 0.979 and an overall RMSE of 0.135 mm/day. Notably, the large model also accurately captures extreme precipitation events, with RMSE values of 0.365 mm/day at the 99.7th percentile and 0.525 mm/day at the 99.99th percentile. All RMSE values for precipitation are computed in log-transformed space using $\log(x + 1)$, where $x$ denotes daily precipitation in millimeters.

\begin{figure*}[t]
\centering
\includegraphics[width=.60\linewidth]{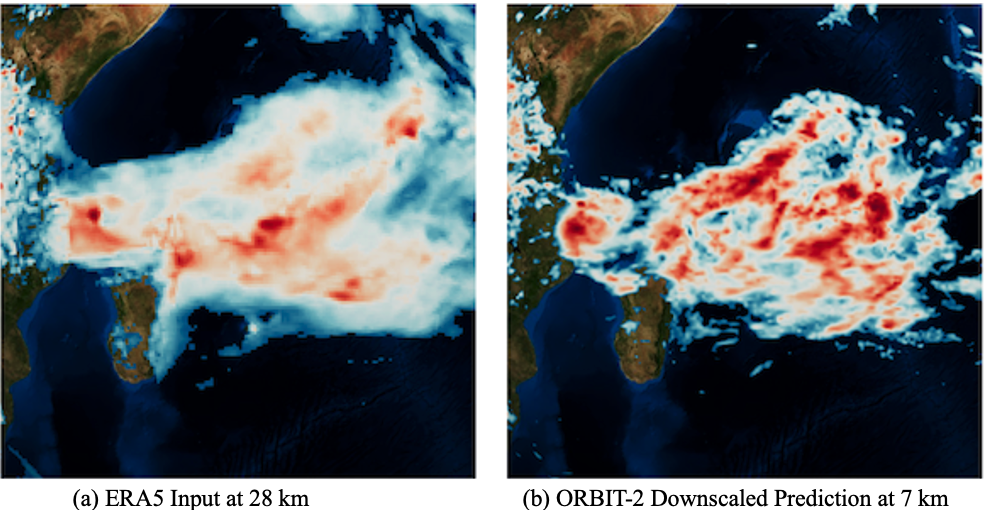}
\vspace*{-0.2cm}
\caption{Example region for global precipitation downscaling during fine-tuning. (a) ERA5 input at 28 km resolution; (b) ORBIT-2 downscaled output at 7 km. Click on the link here for an interactive visualization of global precipitation downscaling for the year 2020 \textcolor{blue}{\href{https://youtu.be/Iahsl1L_1jQ}{Click Here}.}
}
\vspace*{-0.4cm}
\label{fig:animation}
\end{figure*}

\begin{figure*}[t]
\centering
\includegraphics[width=.9\linewidth]{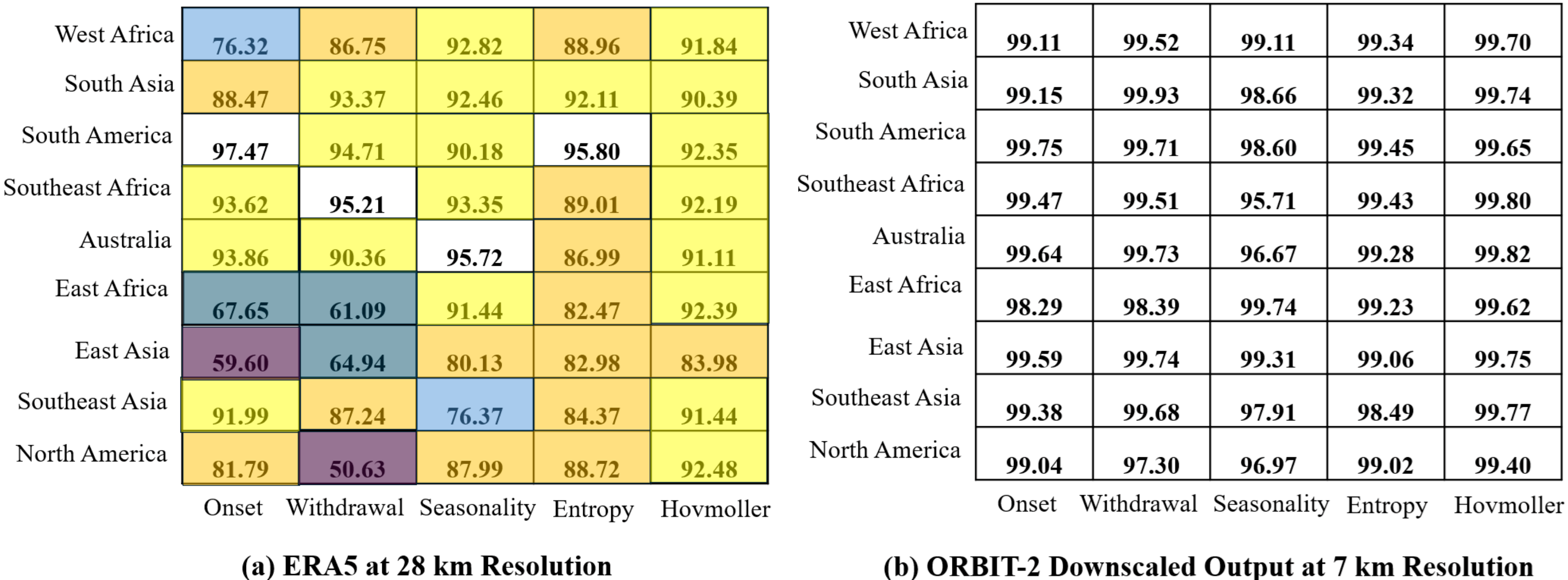}
\vspace*{-0.2cm}
\caption{(a) ERA5 skill scores for global precipitation at 28 km resolution, evaluated using averaged $R^2$ and SSIM, over 24 years (1998-2021). Analysis metrics include monsoon onset and withdrawal, seasonality, entropy, and Hovmöller diagnostics. (b) ORBIT-2 skill scores for global precipitation at 7 km resolution, showing improved fidelity across the same metrics.}
\vspace*{-0.4cm}
\label{fig:region_by_region_downscaling_accuracy}
\end{figure*}

\noindent\ul{\textbf{Global Precipitation Fine-Tuning.}}
ORBIT-2's fine-tuning performance for global precipitation downscaling task is summarized in Table~\ref{tab:merged_finetuning} row 4. Both model sizes (9.5M and 126M parameters) achieve similar accuracy, with $R^2 = 0.98$, SSIM = 0.92, PSNR = 35.0, and RMSE = 0.09 mm/day (in $\log(x + 1)$ space). These results underscore ORBIT-2’s robustness across both regional and global downscaling tasks.

For visualization, Fig.~\ref{fig:animation} shows a one-day snapshot of an example region, comparing  ERA5 precipitation at 28 km with ORBIT-2 downscaling at 7 km. A full-year animation is available online \textcolor{blue}{\href{https://youtu.be/Iahsl1L_1jQ}{Click Here}}, illustrating the spatial resolution gains and improve climate pattern for year 2020.

To evaluate long-term fidelity and climate pattern, Fig.~\ref{fig:region_by_region_downscaling_accuracy} reports performance of high-resolution downscaling in capturing key precipitation characteristics for 1998–2021 across 58 climatically homogeneous land and ocean regions defined by the Intergovernmental Panel on Climate Change (IPCC). For each region, we report a composite skill score, defined as the averaged $R^2$ correlation and SSIM (scaled between 0 to 100), with higher values indicating stronger agreement with the IMERG 7 km observational reference.

Performance is further analyzed in nine monsoon regions, where precipitation is governed by strong seasonality and complex land ocean atmosphere interactions, and significant hydrological and socio-economic impacts. Following~\cite{ashfaq2021robust}, we evaluate monsoon onset and withdrawal timing, precipitation seasonality and entropy, and the northward progression of rainfall (via Hovmöller diagnostics).
Results show substantial improvements with ORBIT-2. As shown in Fig.\ref{fig:region_by_region_downscaling_accuracy}(a), ERA5 at 28 km resolution exhibits moderate skill scores across metrics and regions, whereas ORBIT-2 7 km downscaling  (Fig.\ref{fig:region_by_region_downscaling_accuracy}(b)) consistently achieves much higher skill scores across all metrics and monsoon regions. These results highlight the effectiveness of ORBIT-2 in enhancing the spatiotemporal fidelity of precipitation, especially in regions governed by complex climatic processes.

\section{Implication}
\label{sec:implication}

ORBIT-2 represents a significant leap forward in the convergence of artificial intelligence, HPC, and Earth system science. By overcoming fundamental challenges in scalability, resolution, and uncertainty, ORBIT-2 sets a new standard for climate and earth system foundation models.

\ul{Impact on HPC}.
ORBIT-2 pushes the frontiers of HPC by enabling ViTs at unprecedented scale. Through its novel TILES algorithm, ORBIT-2 reduces the self-attention complexity from quadratic to linear, allowing for efficient processing of ultra-long sequences. Complementing this, the Reslim architecture introduces a lightweight, uncertainty-aware learning framework that leverages residual learning and Bayesian regularization to improve efficiency and training robustness.

Crucially, ORBIT-2 achieves breakthrough scalability in ViT training, scaling up to 10 billion parameters model size across 65,536 GPUs, and scale up to 4.2 billion token sequence length, several magnitudes longer than the state-of-the-art long sequence implementation that scales to 188K tokens. It sustains up to 4.1 ExaFLOPs of performance with 74–98\% strong scaling efficiency at 65,536 GPUs, setting a new benchmark for exascale AI workloads.  ORBIT-2 serves as a blueprint for next-generation exascale foundation models, enabling transformative applications across domains including genomics, fluid dynamics, astrophysics, and Earth system modeling.

\ul{Impact on Climate Science}.
ORBIT-2 enables hyper-resolution, global-scale downscaling with state-of-the-art accuracy and efficiency. Against observations, it achieves an $R^2$ of 0.999 for temperature and 0.979/0.986 for precipitation at 7 km resolution over the continental United States and globally, respectively. Unlike traditional approaches, ORBIT-2 generalizes across variables and regions using a single foundation model, improving climate projection fidelity, resolving fine-scale processes, and detecting localized extremes. This unified capability makes ORBIT-2 a valuable tool for supporting climate adaptation and mitigation strategies. Besides the above capability, this paper also shows ORBIT-2 foundation AI model's capability to perform global downscaling in 4 millisecond for 9.5M parameter model and 0.5 second for a 10B parameter model with only 8 GPUs through model inferencing, showing the AI model's capability for near-real-time climate and weather prediction on edge devices. This is transformative compared to the conventional numerical simulation methods, which often takes hours and days to compute on a large supercomputer.

\ul{Limitations and Practical Considerations}.
Like all AI-based methods, ORBIT-2 may inherit biases from training data source and lacks strict enforcement of physical conservation laws, requiring caution in applications sensitive to energy or mass balance. Practical deployment will require careful consideration on application-based bias correction, physics-informed integration, and domain-specific fine-tuning to ensure reliability across diverse climates and regimes. Finally, while near-real-time inference on limited hardware is promising, robust evaluation in operational settings is needed before widespread deployment in climate services and decision-making.

\section*{Acknowledgments}
We thank Jonathan Coles for helping us get scalability numbers on the Alps cluster. This manuscript has been authored by UT-Battelle, LLC, under contract DE-AC05-00OR22725 with the US Department of Energy (DOE). The U.S. Government retains a nonexclusive, worldwide license to publish or reproduce the published form of this manuscript, or to authorize others to do so, for U.S. Government purposes, as acknowledged by the publisher.
This research was primary supported by the ORNL's AI Initiative sponsored by the Director's Research and Development Program at ORNL, additionally supported by DOE Early Career Project sponsored by the BER program. It was also supported as part of the Energy Exascale Earth System Model (E3SM) project, funded by the U.S. Department of Energy Office of Science, Office of Biological and Environmental Research, and Earth Systems Model Development Program.  An award of computer time was provided by the INCITE program through the Oak Ridge Leadership Computing Facility, which is a DOE Office of Science User Facility.
\bibliographystyle{ACM-Reference-Format}
\bibliography{main}

\end{document}